\documentclass{article}
\usepackage[preprint]{icml2026}
\usepackage{booktabs}
\usepackage{multirow}
\usepackage{tikz}
\usepackage{amsmath}
\usepackage[binary-units]{siunitx}
\usepackage[abbreviations]{foreign}
\usepackage{xspace}
\usepackage[inline]{enumitem}
\usepackage[algo2e,ruled,lined,boxed,commentsnumbered, noend, linesnumbered, procnumbered]{algorithm2e}
\usepackage{hyperref}

\usepackage{pifont}%
\usepackage{ifthen}
\newboolean{showcomments}
\setboolean{showcomments}{false}
\ifthenelse{\boolean{showcomments}}
{ \newcommand{\mynote}[3]{
		\fbox{\bfseries\sffamily\scriptsize#1}
		{\small$\blacktriangleright$\textsf{\emph{\color{#3}{#2}}}$\blacktriangleleft$}}
	\newcommand{\zzz}[1]{{\setlength{\fboxsep}{2pt}\fcolorbox{black}{yellow}{\textsf{\emph{#1}}}}\xspace}}
{ \newcommand{\mynote}[3]{}
	\newcommand{\zzz}[1]{}}

\usepackage{acronym}
\acrodef{DL}{decentralized learning}
\acrodef{ML}{machine learning}
\acrodef{D-PSGD}{decentralized parallel stochastic gradient descent}
\acrodef{FL}{federated learning}
\acrodef{SGD}{stochastic gradient descent}
\acrodef{IID}{independent and identically distributed}
\acrodef{non-IID}{non independent and identically distributed}
\acrodef{RMSE}{root mean square error}
\acrodef{RMW}{random model walk}
\acrodef{GL}{gossip learning}
\acrodef{EL}{epidemic learning}
\acrodef{DWT}{discrete wavelet transform}
\acrodef{FFT}{fast Fourier transform}
\acrodef{MI}{mutual information}
\acrodef{DP}{differential privacy}
\acrodef{VN}{virtual node}
\acrodef{RN}{real node}
\acrodef{LDP}{local differential privacy}
\acrodef{PNDP}{pairwise network differential privacy}
\acrodef{PNLDP}{pairwise network local differential privacy}
\acrodef{GI}{gradient inversion}
\acrodef{CML}{collaborative machine learning}
\acrodef{TPR}{true positive rate}
\acrodef{FPR}{false positive rate}

\acrodef{LA}{linkability attack}
\acrodef{GIA}{gradient inversion attack}
\acrodef{MIA}{membership inference attack}
\acrodef{AIA}{attribute inference attack}
\acrodef{ROC}{receiver operating characteristic}
\acrodef{AUC}{area under the ROC curve}

\acrodef{LLM}{large language model} %
\newcommand{\sys}{\textsc{AWO}\xspace}

\newcommand{\react}{{\xspace}ReAct\xspace}
\newcommand{\appworld}{{\xspace}\textsc{AppWorld}\xspace}
\newcommand{\visualwebarena}{{\xspace}\textsc{VisualWebArena}\xspace} %
\graphicspath{ {figures/} }
\usepackage{tikz}
\usepackage[eulergreek]{sansmath}
\usepackage{pgfplots}
\usepackage{pgfplotstable}
\usepackage{cleveref}
\usepackage{comment}
\crefname{assumption}{assumption}{assumptions}

\pgfplotsset{compat=newest}
\usepgfplotslibrary{external,units,colorbrewer,groupplots,fillbetween,statistics}
\tikzsetexternalprefix{figures/}
\tikzset{external/mode=list and make}
\usetikzlibrary{patterns,shapes.misc}

\newcommand{\newgroupwidth}[2]%
{\expandafter\xdef\csname groupwidth#1\endcsname{#2}}

\newcounter{groupwidth}
\newsavebox{\groupwidthbox}
\makeatletter
{\edef\groupnumber{#1}%
	\stepcounter{groupwidth}%
	\@ifundefined{groupwidth\thegroupwidth}{\pgfmathsetlengthmacro{\mywidth}{\linewidth/\groupnumber}}%
	{\expandafter\let\expandafter\mywidth\csname groupwidth\thegroupwidth\endcsname}%
	\begin{lrbox}{\groupwidthbox}%
		\tikzset{/pgfplots/width={\mywidth}}%
		\ignorespaces}%
	{\end{lrbox}%
	\usebox\groupwidthbox
	\pgfmathsetlengthmacro{\mywidth}{\mywidth + (\linewidth - \wd\groupwidthbox)/\groupnumber}
	\immediate\write\@auxout{\string\newgroupwidth{\thegroupwidth}{\mywidth}}}
\makeatother
\usepackage{amsmath,amssymb,amsfonts,amsthm}
\usepackage{physics}
\usepackage{enumitem}
\usepackage{thm-restate}
\usepackage{bbm}
\theoremstyle{definition}

\theoremstyle{remark}

\allowdisplaybreaks

 \usepackage[most]{tcolorbox}
\usepackage{fancyvrb}

\begin{document}

\twocolumn[

  \icmltitle{Optimizing Agentic Workflows using Meta-tools}
%
%
  %
  %
  %

  %
  %
  %
  %

  %
  %
  %

  \begin{icmlauthorlist}
    \icmlauthor{Sami Abuzakuk}{sch}
    \icmlauthor{Anne-Marie Kermarrec}{sch}
    \icmlauthor{Rishi Sharma}{sch}
    \icmlauthor{Rasmus Moorits Veski}{sch}
    \icmlauthor{Martijn de Vos}{sch}
  \end{icmlauthorlist}

  \icmlaffiliation{sch}{EPFL, Switzerland}

  \icmlcorrespondingauthor{Sami Abuzakuk}{sami.abuzakuk@epfl.ch}
  \icmlcorrespondingauthor{Rasmus Moorits Veski}{rasmus.veski@epfl.ch}

  \icmlkeywords{Machine Learning, ICML}

  \vskip 0.3in
]

\printAffiliationsAndNotice{}

\begin{abstract}

Agentic AI enables \acfp{LLM} to dynamically reason, plan, and interact with tools to solve complex tasks.
However, agentic workflows often require many iterative reasoning steps and tool invocations, leading to significant operational expense, end-to-end latency and failures due to hallucinations.
This work introduces Agent Workflow Optimization (\sys), a framework that identifies and optimizes redundant tool execution patterns to improve the efficiency and robustness of agentic workflows.
\sys analyzes existing workflow traces to discover recurring sequences of tool calls and transforms them into \emph{meta-tools}, which are deterministic, composite tools that bundle multiple agent actions into a single invocation.
Meta-tools bypass unnecessary intermediate \ac{LLM} reasoning steps and reduce operational cost while also shortening execution paths, leading to fewer failures.
Experiments on two agentic AI benchmarks show that \sys reduces the number of \ac{LLM} calls up to 11.9\% while also increasing the task success rate by up to 4.2 percent points.

\end{abstract}

\section{Introduction}
\label{sec:intro}

Agentic AI has emerged as a powerful paradigm for solving complex, open-ended tasks with minimal human intervention~\cite{acharya2025agentic}.
By combining \acfp{LLM} with the ability to call tools, store memory, and perform iterative reasoning, agentic systems can dynamically plan and adapt while interacting with external environments.
This flexibility allows \ac{LLM} agents to execute a wide variety of workflows and solve user tasks, \eg, making bookings across multiple services or managing personal finances, without requiring developers to explicitly program deterministic control flows in advance.
Because of their flexibility, \ac{LLM} agents are being integrated across an increasing number of domains, such as software engineering~\cite{he2025llm}, web-based task automation~\cite{ning2025survey}, and customer support~\cite{wang2025ecom}.

To enable such flexibility, \ac{LLM} agents can access a set of pre-defined \emph{tools}, each exposing a specific capability of the system.
For example, developers can build tools to send emails, write to files, or search the web.
In practice, these tools typically correspond to existing API endpoints of services that allow agents to retrieve information, modify state, or interact with external services~\cite{schick2023toolformer,ning2025survey,wu2024copilot}.
To solve a task, the \ac{LLM} outputs code that invokes particular tools with the required parameters.

However, deploying \ac{LLM} agents comes at a significant operational cost, introducing substantial overhead in both latency and resource consumption.
Tool-calling agents invoke an \ac{LLM} multiple times per task to plan, reason, revise, and recover from intermediate failures~\cite{yao2023react}.
Each reasoning step incurs inference cost and contributes to end-to-end latency, which can be prohibitive in user-facing applications.
This is particularly true if solving a task requires the execution of many tool calls in sequence, as is the case when interacting with fine-grained APIs.
For example, creating a Spotify playlist and populating it using the web API requires multiple sequential requests (authorization, playlist creation, and playlist item addition).
While agentic systems that work with such fine-grained tools end up being more flexible, the inherent nondeterminism of \acp{LLM} can lead agents to explore redundant or suboptimal actions, particularly in multi-step tasks, further increasing computational costs and execution time.
As agentic systems are deployed in increasingly complex environments and integrated into production workflows, these costs become a primary barrier to adoption, scalability and reliability~\cite{zhangcut}.

\begin{figure}[t]
    \centering
    \includegraphics[width=.8\linewidth]{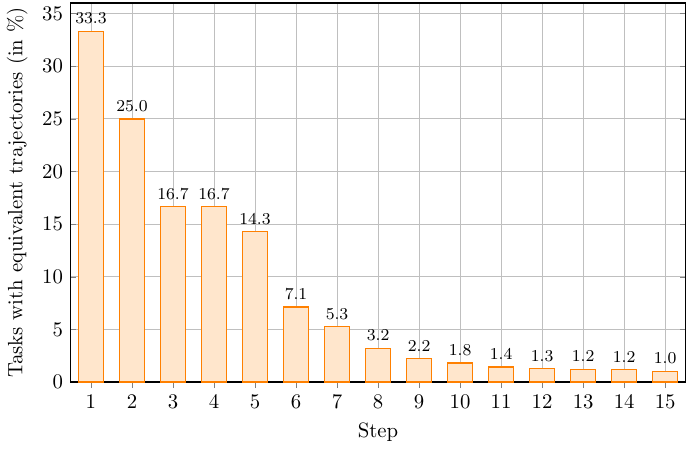}
    \caption{Percentage of tasks where agents have equivalent trajectories at a given step in \appworld (lower bound).}
    \label{fig:tool_call_duplication}
\end{figure}

Examining common agentic workflows, we observed that they often exhibit highly regular structure, particularly in their early stages.
We executed the tasks in \appworld, a public agentic AI benchmark~\cite{trivedi2024appworld}, and captured the trajectories that the agents follow.
In \Cref{fig:tool_call_duplication}, we present the lower bound of the number of tasks where the agents follow the same trajectory per step.
For example, after 5 steps, at least 14.3\% of the tasks being completed by the agent followed an equivalent trajectory.
We attribute this duplication to the fact that the set of exposed tools is typically not designed with agentic usage in mind.
Therefore, identifying and deduplicating LLM effort across tasks has significant potential in reducing the operational overhead of running agentic systems.

In this paper, we introduce \emph{Agent Workflow Optimization \sys}, a generic framework for improving the efficiency, utility, and runtime cost of agentic systems.
\sys works by identifying and coalescing structural tool calls in trajectories, directly reducing the number of LLM calls in the agentic workflows.
\sys provides the means to analyze tool call trajectories and detect recurring sequences of tool invocations that correspond to routine sub-tasks, and compiles these sequences into \emph{meta-tools}.
These meta-tools encapsulate common multi-step behaviors into single callable operations, allowing agents to bypass intermediate reasoning while retaining the ability to flexibly compose workflows at a higher level.
In this way, \sys reduces inference cost and latency without sacrificing the generality that makes agentic systems effective.

We evaluate the effectiveness of \sys using two representative agentic AI benchmarks, covering both API-driven application workflows and interactive web-based tasks.
Across these settings, we analyze large collections of execution traces generated by state-of-the-art \ac{LLM} agents and identify recurring tool invocation patterns that can be compiled into meta-tools.
Our results show that integrating meta-tools substantially reduces the number of \ac{LLM} reasoning steps and tool calls required to complete tasks, yielding reductions of \ac{LLM} calls up to 11.9\%.
In addition, we observe that \sys improves the task success rates by up to 4.2 percentage points, indicating that reducing redundancy not only improves system's efficiency but can also enhance the effectiveness of the agents.

In summary, our contributions are as follows:
\begin{itemize}
    \item We introduce \sys, a novel framework for detecting redundant patterns in an agentic workflow through merging agentic executions (\Cref{sec:design}). This allows for the creation of \emph{meta-tools} removing unecessary reasoning steps, reducing the number of tokens used, and hence, costs when operating agentic AI.
    \item We thoroughly evaluate \sys on a diverse set of agentic workflows, demonstrating its effectiveness in reducing token usage up to 11.9\% and improving task success rate by 4.2 percent points (\Cref{sec:results}).
    \item We release \sys as an open-source framework~\footnote{Removed for double-blind review}, enabling reproducibility, community collaboration, and motivating further research advancement in optimization agents.
\end{itemize}

\begin{figure}[b]
    \centering
    \includegraphics[width=\linewidth]{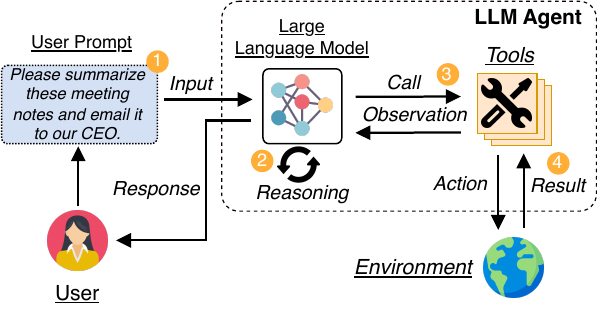}
    \caption{The \react Loop. The agent interleaves reasoning with external execution (``Action''), creating a step-by-step workflow.}
    \label{fig:react_loop}
\end{figure}

\begin{figure*}[t]
    \centering
    \includegraphics[width=\linewidth]{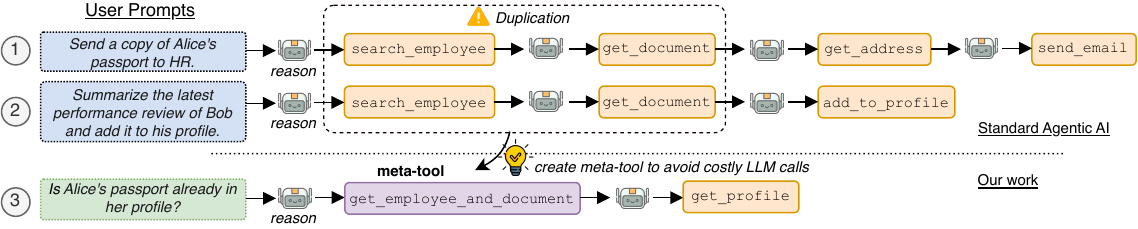}
    \caption{\emph{(top)} Two distinct user prompts (in blue) and the associated workflow, including \ac{LLM} reasoning and tool calling. User prompts, while different from a surface-level, might share sequences of tool calls. Our framework leverages this property to create \emph{meta-tools}, reducing the number of costly \ac{LLM} calls.}
    \label{fig:motivating_example}
\end{figure*}

\section{Background and Motivation}
\label{sec:prelims}

We first describe the technical underpinning of agentic AI, explain workflows and trajectories, and then illustrate using an example how existing workflows can be made more efficient by coalescing tool calls.

\subsection{Agentic AI and \react}
\label{sec:agentic_ai}
Agentic AI refers to systems in which an \ac{LLM} is used as a decision-making component that can autonomously pursue goals through interaction with an external environment.
An agent maintains an internal state, such as task context, intermediate results, or memory, and dynamically selects actions to execute based on this state.
These actions typically take the form of tool invocations, which may query external information sources, modify persistent state, or trigger side effects in downstream systems.

A common realization of agentic AI is the ReAct (Reasoning and Acting) loop, in which an agent interleaves internal reasoning with external actions~\cite{yao2023react}.
In this paradigm, the \ac{LLM} alternates between generating natural-language reasoning steps that reflect its current plan or hypothesis, and invoking tools that act on the environment.
We show the \react loop and components of the \ac{LLM} agent in \Cref{fig:react_loop}.
A user prompt is provided to an \ac{LLM} and the \ac{LLM} reasons about it.
It may then decide to call tools, which modify the environment.
The outcome of each tool invocation is returned to the agent as an observation, which is then incorporated into subsequent reasoning.
This iterative process continues until the agent determines that the task has been completed.
We refer to this continuous cycle as the \emph{agent workflow}.
The \react loop enables agents to decompose complex tasks into smaller steps, recover from intermediate errors, and adapt their strategy based on observations by coupling reasoning with action and feedback.

\subsection{Tools}
\label{sec:tools}

Tools are the primary interface through which an \ac{LLM} agent interacts with its environment.
Conceptually, a tool exposes some capability of the underlying system, such as querying an external service, retrieving information, or modifying a persistent state.
In agentic AI implementations, each tool has a unique name, a natural-language description that informs the LLM on when to invoke it, and a structured schema describing its input arguments and return values.
At runtime, the agent does not execute tools directly.
Instead, the \ac{LLM} produces a structured output indicating the selected tool and the corresponding arguments, which is parsed and executed by the system.
As shown in \Cref{fig:react_loop}, the execution result is then returned to the agent as an observation and incorporated into subsequent reasoning steps.

\subsection{Duplication in Tool Calls Across Prompts}
To motivate our work, we show in \Cref{fig:motivating_example} different user prompts and the corresponding tool calls and \ac{LLM} reasoning steps.
Our example assumes an agentic system in a company that is made available to company employees.
The agentic system has access to a set of tools that enables the agent to search for employees, fetch documents, manage employee profiles, and send emails.
A prompt in this setting could be "Send a copy of Alice's passport to HR".
To solve this task, an \ac{LLM} agent could first call the \textsc{search\_employee} function to obtain the employee identifier of Alice and then get her passport by calling the \textsc{get\_document} function.
To send the passport to HR, the agent can leverage the \textsc{get\_address} and \textsc{send\_email} tools.
Between tool calls, we invoke the \ac{LLM} with either the user prompt or the output of the last tool call, which is a compute-intensive and costly operation.

Another prompt, for example, "Summarize the latest performance review of Bob and add it to his profile" might use initially the same two tools as the first prompt, although with different arguments.
While our example is hypothetical, such tool call duplication actually occurs in real-world agentic workloads.
As highlighted by \Cref{fig:tool_call_duplication}, agents regularly enter equivalent states implying that the agents called identical tools or a set of tools leading to the same consequence.
Therefore, we can identify shared sequences of tool calls across prompts and replace this sequence with a single \emph{meta-tool}.
We show such a meta-tool in purple in \Cref{fig:motivating_example} for a third prompt "Is alice's passport already in her profile?".
This meta-tool, named \textsc{get\_employee\_and\_document}, combines the functionality of the \textsc{search\_employee} and \textsc{get\_document} tool calls and avoids the intermediary \ac{LLM} step.

Based on the above insight, our work focusses on the following research question: \emph{How can we automatically identify and exploit recurring tool-invocation patterns in agentic execution traces in order to reduce reasoning overhead, latency, and cost without sacrificing task generality?}

\begin{figure*}[t]
    \centering
    \includegraphics[width=\linewidth]{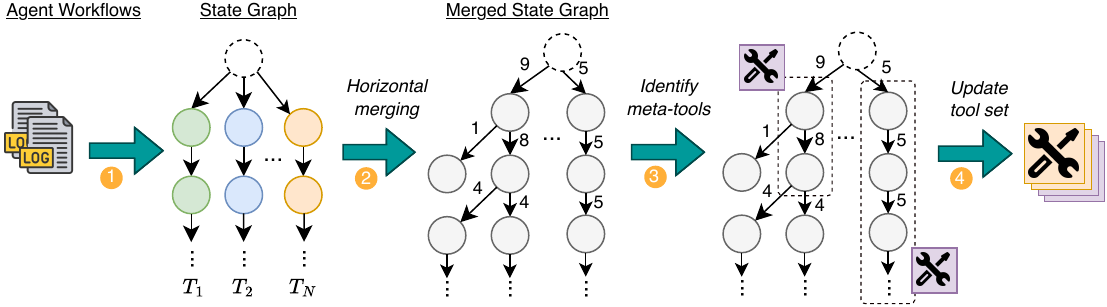}
    \caption{Overview of the \sys framework. Existing agent workflows are first transformed into a state graph (step 1) and then compressed via horizontal merging into a merged state graph to expose redundant workflows that end up in equivalent states (step 2). The merged state graph is used to identify meta-tools (step 3) which are added to the agent’s tool set (step 4).}
    \label{fig:system}
\end{figure*}

\section{Design of \sys}
\label{sec:design}

We now describe the workflow of the \sys framework, which steps are visualized in \Cref{fig:system}.
Agents frequently exhibit convergent patterns where the they repeat identical sequences of thoughts and tool calls to solve recurrent sub-problems.
Essentially, \sys first analyses existing agentic workflows to identify these redundant sub-sequences and then combines them into meta-tools, thereby removing the need for \ac{LLM} inference in those sections.
\sys assumes the representativeness of the existing workflows, as the optimization will work only on sufficiently similar future workflows.
Our approach increases the efficiency and robustness of agentic workflows while not compromising the flexibility of the \ac{LLM} agent.
We first describe the high-level process of \sys in \Cref{sec:nutshell} and then describe the workflow more formally in the subsequent subsections.

\subsection{\sys in a Nutshell}
\label{sec:nutshell}

To discover potential locations for meta-tools in a system we map existing agentic executions into a \emph{state graph} and apply graph merging to merge similar executions (step 1 in \Cref{fig:system}).
In a state graph, each node (except for the root node) indicates the system state and each edge denotes a state transition, \eg, induced by a \ac{LLM} call.
Each path from the root node to a leaf node is a single execution, which are identified by different colors in \Cref{fig:system}.
We first apply \emph{horizontal merging} that aims to identify and combine states across executions, resulting in a \emph{merged state graph} (step 2).
The motivation behind this is that different sequences of tool calls can end up in the same state and we can use it to identify commonly executed paths.
For example, it does not matter in which order an agent interacts with a calendar event delete action, if the same events were deleted.
Horizontal merging is an application-specific process and requires domain knowledge.
In this new graph, each edge is weighted, indicating the number of executions that follow that edge.
Next, we apply \emph{vertical merging}, which identifies the meta-tools with the highest potential to reduce \ac{LLM} calls (step 3).
We then select a subset of these identified meta-tools and add them to the existing set of tools that the \ac{LLM} agents can access (step 4).

In the remainder of this section, we more formally describe the \sys workflow and involved steps.

\subsection{State Graphs}
\label{sec:state_graphs}

To discover recurrent patterns, we need to have a representation of all executions in one data structure.
In the context of \sys, an agentic execution $E_i$ is defined as an ordered sequence of tool calls:
\[
E_i = (T_{i}^{1}, T_{i}^{2}, \ldots, T_{i}^{n_i}),
\]
where $T_{i}^{j}$ denotes the $j$-th tool call in execution $i$, and $n_i$ is the length of execution $i$.
Note that we do not model intermediate \ac{LLM} reasoning or generation steps as part of the execution, as they are internal, non-deterministic, and not externally observable, whereas tool calls constitute the agent’s explicit interaction with the environment.
We then define a state graph $ S = (V, E) $ as a graph, where edges can be interpreted as the decision to go from one tool call to the other, and node as as the history (list) of the agent's tool calls so far.
For agentic workflows, the decision of which tool to call is determined by the output of the agents' \ac{LLM} call at that step.
In \Cref{fig:system}, we show an example state graph with $N$ executions.

\subsection{Merging the State Graph}
\label{sec:compressing_state_graphs}

Horizontal merging targets redundancy across agentic executions by identifying common tool-calls that repeatedly occur in different runs.
Given a collection of executions $\{E_i\}$, we introduce domain expert's rules to merge semantically equivalent states such that identical steps are represented by the same node.
For example, tools that read data are typically commutative and can be applied in any order.
A domain expert rule would make such actions commutative.
We remark that identification of mergeable prefixes is highly application-dependent: determining which tool-call sequences are semantically equivalent and safe to reuse requires domain knowledge about tool behavior, side effects, and return values.
As a result, it is difficult to effectively automate horizontal merging and \sys relies on domain experts to specify valid merging boundaries and equivalence criteria.
However, we experiment with using \ac{LLM} agents for horizontal merging, which is described in Appendix~\ref{sec:automated_loop}.

The output of the horizontal merging process is a \emph{merged state graph} $S_m = (V, E, w) $.
Note that $ S_m $ contains weighted edges with weight $w$, where the weight of an edge between two states $ s_1 $ and $ s_2 $ indicates the number of executions in which a transition from $s_1$ to $s_2$ occurs.
An example merged state graph is visualized in \Cref{fig:system}.

\subsection{Identifying Meta-tools}

We now discuss how we extract meta-tools from a merged state graph $S_m$ and threshold $T$, which defines the minimal amount of executions that are worth merging. %
This algorithm is shown in \Cref{alg:metatool}.
We operate on a mutable copy of $G$, referred to as $G'$ (Line 1 in \Cref{alg:metatool}).
The main idea is that we iteratively select some node $n$ in $G'$ and greedily attempt to merge these nodes with one of the children of $n$ as long as this reduces the total sum of weights on all edges in $G'$, \eg, the number of \ac{LLM} calls.
We thus successively accumulate meta-tools $M$ and recompress $G_m$ into $G'_m $ once a meta-tool is identified.
This process repeats until we cannot identify any more meta-tools, after which we add them to the toolbox of the \ac{LLM} agent.
In other words, we are not reducing the dynamicity of the agent, but, we are introducing additional static and deterministic routes in the trajectory that the agent can follow.
We now describe \Cref{alg:metatool} in detail.

\begin{algorithm2e}
\caption{Meta-Tool extraction from a merged state graph}
\label{alg:metatool}
\KwIn{Merged state graph $G_m = (V, E, w)$, threshold $T$}
\KwOut{Set of meta‑tools $M$}
$G_m' \leftarrow G_m$\;
\While{true}{
    $\textit{state\_pairs} \leftarrow \textsc{extract\_state\_pairs}(G'_m, T)$\;
    \If{\textbf{not} \textit{state\_pairs}.\textsc{empty}()}{
        $(n_x, n_y) \leftarrow \textit{state\_pairs}[0]$\;
        $\textit{candidate\_tool} \leftarrow \textit{candidate\_tool} \cup \{n_x, n_y\}$\;
        \While {$n_z$ = \textsc{select\_child}($n_y$, $T$)} {
            $\textit{candidate\_tool} \leftarrow \textit{candidate\_tool} \cup \{n_z\}$\;
            $n_y \leftarrow n_z$\;
        }

        $M \leftarrow M \cup \{\textit{candidate\_tool}\}$\;
        $G_m' \leftarrow \textsc{compress\_graph($G_m$, $M$)}$\;
        $\textit{candidate\_tool} \leftarrow \emptyset$\;
    }
    \Else{
        \KwRet{$M$}
    }
}
\end{algorithm2e}

\textbf{State Pair Extraction (Lines 2--3).}
We start the identification of a meta-tool by extracting all pairs of edges $n_x \to n_y$ in $G'$, filtering those with $w(n_x,n_y) \geq T$ (line 2).
$T$ is a hyperparameter of \sys and ensures that the identified meta-tools sufficiently reduce the number of \ac{LLM} calls as otherwise we would could end up with a single meta-tool per trajectory.
State pair extraction is handled by the \textsc{extract\_state\_pairs} function that returns an ordered list of node pairs that are connected by an edge.
This list is ordered descendingly by weight and then ascendingly by the path length to the root as tiebreaker.

\textbf{Pair and Chain Selection (Lines 5--9)}
If there are no eligible state pairs, the algorithm terminates and returns the set of currently identified meta-tools $M$ (line 14).
Otherwise, we select the first pair $(n_x,n_y)$ from the extracted state pairs (line 5).
This edge constitutes the starting point of our candidate meta-tool that we refer to as \emph{candidate\_tool} (line 6).
Then, we recursively extend the meta-tool by selecting an edge from the children of $n_y$.
A child will be selected by the \textsc{select\_child} function only if it satisfies:
\[
w(n_y, n_z) > \frac{1}{2} \sum_{y' \in N^{+}(n_y)} w(n_y, y')
\]
where $N^{+}(n_y)$ is the set of children of $n_y$.
Intuitively, merging $n_x, n_y, n_z$ would require to keep nodes $n_x$ and $n_y$ for all the children of $n_y$ except $n_z$.
If the number of trajectories covered by these nodes are more than $50\%$ of the trajectories using $n_z$, the compression would increase the global number of edges.
We extend the candidate meta-tool until no more children can be selected (lines 7--9).

\textbf{Rule Accumulation and Compression (Lines 10--14)}
If we cannot extend the candidate meta-tool anymore, we add the set of edges generated by this iteration to the set $M$ (line 10).
For example, a candidate meta-tool of $\{n_x,n_y,n_z\}$ represents one meta-tool calling the tools transitionning from state $n_x$ to $n_y$ and $n_y$ to $n_z$ in sequence.
We then compress the original graph $G_m$ using all accumulated rules in $M$, merging matched sequences into new nodes (line 11).
We repeat until no valid state pairs remain in $G_m'$.
Finally, the algorithm outputs $M$ which contains the list of meta-tools that should be created for this set of trajectories.
\section{Experimental Setup}
\label{sec:exp_setup}

We implemented our agents following the \react framework (see \Cref{sec:agentic_ai}) and run the benchmarks using the \textsc{GPT 5.1}~\cite{OpenAI_GPT51_2025} and \textsc{Claude-sonnet-4-5-20250929}~\cite{anthropic2025claude45sonnet} models.%
For all models, we fix the temperature to $1$. For generality, we also ran the experiments on an open sorce model \textsc{GPT-OSS 120B}~\cite{OpenAI_GPTOSS_2025}, the results are found in Appendix~\ref{app:appendix_results}.
We next describe the benchmarks, what metrics we use to validate our results and describe the process of extracting meta-tools from these specific benchmarks.

\subsection{Benchmarks and meta-tool identification}
We evaluate the effectiveness of \sys on two benchmarks for agentic AI: \visualwebarena~\cite{koh2024visualwebarena} and \appworld~\cite{trivedi2024appworld}.
These benchmarks are well-suited for our setting as they require multi-step tool use across heterogeneous APIs, producing long agentic trajectories.
We now describe the specifications and the \sys meta-tool identification process of these benchmarks.

\textbf{\visualwebarena.}
\visualwebarena~\cite{koh2024visualwebarena} is a multimodal benchmark that simulates real-world web applications.
The benchmark comprises \num{910} tasks using three websites: Reddit (210 tasks), Classifieds (234 tasks) and Shopping (466 tasks).
The tools included in this benchmarks allow the agent to interact with the website, \eg, click on page elements, type text in an input field, or change tab.
Our evaluation utilizes the tool-calling agent included in the benchmark.
An example task in \visualwebarena in the Shopping category is: \emph{Buy the cheapest color photo printer and send it to Emily's place}.

Analyzing agent traces from \visualwebarena, we developed domain-aware merging rules to abstract ephemeral UI details such as element IDs, URLs and repeated navigation steps, while preserving semantic intent.
We identified two meta-tools for each task category to capture recurring patterns like search-and-navigate flows and form submissions.
Further details are provided in Appendix~\ref{app:vwa_metatools}.

\textbf{\appworld.}
\appworld~\cite{trivedi2024appworld} is a benchmark simulating nine real-world applications such as Gmail and Spotify.
These applications are operable via \num{457} distinct APIs, populated with simulated data of 100 fictitious users.
Our analysis was conducted on all \num{168} tasks in the \textsc{test-normal} split.
For this benchmark, we use the coding agent configuration, as this was the best-performing agent reported in the original benchmark paper.
This means that when merging tools we no longer explicitly reduce LLM calls, but do it indirectly through giving the LLM more direct tools.
An example task in \appworld is: \emph{Like all the songs and albums in my Spotify song and album library, respectively, that I have not liked yet.} %

To identify meta-tools in \appworld, we analyzed execution traces generated across the full benchmark and constructed a state graph capturing tool invocation sequences.
Horizontal merging was guided by domain knowledge about API semantics and their side effects.
In particular, we treated read-only operations as commutative, collapsed repeated authentication flows, and abstracted away user-specific identifiers via regex-based normalization.
This process revealed a small set of highly dominant prefixes that appeared across nearly all successful executions.
We identified five meta-tools, all corresponding to application-specific auto-login and session-initialization routines.
We provide additional details about merging heuristics and meta-tool definitions in Appendix~\ref{app:appworld_metatools}.

\subsection{Metrics}
\label{sec:metrics}

Our primary evaluation objective is to quantify the efficiency gains achieved by \sys.
Thus, the main focus is on reducing the \emph{number of LLM calls} per benchmark execution, as each call directly incurs inference cost, contributes to end-to-end latency and expends the \ac{LLM} context, increasing the likelyhood of hallucinations.
We also report the resulting \emph{token usage} and \emph{monetary cost} of \ac{LLM} invocations, with and without \sys, which are derived from the total number of input and output tokens processed.

To provide further insight into how workflow optimization affects the behavior of \ac{LLM} agents, we also report execution-level characteristics.
We measure the \emph{length of execution trajectories}, expressed as the number of reasoning or action steps per task, which captures the extent to which meta-tools shorten agent workflows.
We also report the \emph{task success rate} to ensure that efficiency improvements do not come at the expense of solution quality and to assess whether reducing early-stage redundancy can improve robustness.
Unless stated otherwise, all metrics are aggregated over all task executions and reported as averages across the benchmark suite.

\begin{table}[t]
    \centering
    \caption{Total token usage and cost of running both benchmarks on the GPT 5.1 and Claude 4.5 models, with and without \sys. A detailed breakdown is provided in the Appendix~\ref{app:appendix_results} in Tables~\ref{tab:token_cost_embedded_vwa} and ~\ref{tab:token_cost_embedded_aw}.}
    \small
    \setlength{\tabcolsep}{5pt}
    \begin{tabular}{l|cc|cc}
        \toprule
        \multirow{2}{*}{\textbf{Metric}} & \multicolumn{2}{c|}{\textbf{GPT 5.1}} & \multicolumn{2}{c}{\textbf{Claude 4.5}} \\
         & Base & with \sys & Base & with \sys \\
        \midrule
        \multicolumn{5}{l}{\visualwebarena} \\
        Total Tokens
            & 95.7M
            & \begin{tabular}{@{}c@{}}90.3M\\(-5.6\%)\end{tabular}
            & 120.2M
            & \begin{tabular}{@{}c@{}}107.9M\\(-10.2\%)\end{tabular} \\
        Total Cost (\$)
            & 44.5
            & \begin{tabular}{@{}c@{}}42.0\\(-5.7\%)\end{tabular}
            & 272.3
            & \begin{tabular}{@{}c@{}}244.5\\(-10.2\%)\end{tabular} \\
        \midrule
        \multicolumn{5}{l}{\appworld} \\
        Total Tokens
            & 34.1M
            & \begin{tabular}{@{}c@{}}29.0M\\(-14.9\%)\end{tabular}
            & 29.8M
            & \begin{tabular}{@{}c@{}}27.0M\\(-9.4\%)\end{tabular} \\
        Total Cost (\$)
            & 31.59
            & \begin{tabular}{@{}c@{}}26.85\\(-15.0\%)\end{tabular}
            & 29.63
            & \begin{tabular}{@{}c@{}}28.39\\(-4.2\%)\end{tabular} \\
        \bottomrule
    \end{tabular}

    \label{tab:token_usage_and_cost}
\end{table}

\section{Evaluation}
\label{sec:results}
We now present and analyze the gains and characteristics of \sys across both benchmarks.

\subsection{Efficiency Gains by \sys}
We first quantify the efficiency gains by \sys, in terms of \ac{LLM} call count, token usage and cost.

\begin{figure}[t]
    \centering
    \includegraphics[width=0.9\linewidth]{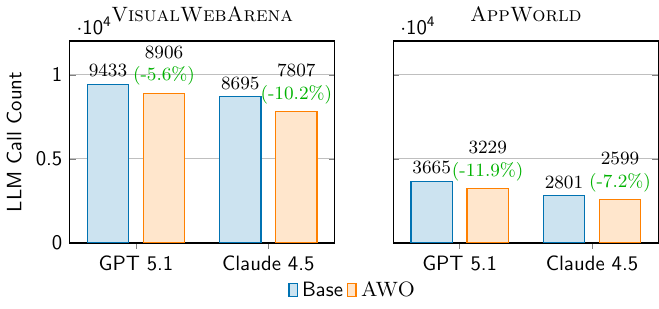}
    \caption{The \ac{LLM} call count with and without \sys, when using the GPT 5.1 and Claude 4.5 models, for \appworld (left) and \visualwebarena (right).}
    \label{fig:llm_call_count}
\end{figure}

\textbf{LLM Calls.}
\Cref{fig:llm_call_count} shows the \ac{LLM} call count with and without \sys, for two models and on both benchmarks.
The number of \ac{LLM} calls made by the agent decreased across both benchmarks when meta-tools are available.
This reduction ranges from 5.6\% to 10.2\% for \visualwebarena and from 7.2\% to 11.9\% for \appworld depending on the model.
Meta-tools thus enable the agent to complete tasks with less reasoning steps required.
Notably, for \visualwebarena we achieve this reduction with only two meta-tools, highlighting the potential for optimization with \sys.

\textbf{Token usage and cost.}
\Cref{tab:token_usage_and_cost} provides an overview of the total (input and output) tokens and cost when running both benchmarks using the evaluated models, with and without \sys.
We observe that the cost savings are directly linked to the llm calls count.
For instance, the reduction ranges from 5.7\% to 10.2\% for \visualwebarena and from 4.2\% to 15\% in \appworld. This is due to the most expensive part of an LLM's work being linked to the cost of its output tokens (a full breakdown of the cost is available in Appendix~\ref{app:appendix_results}).
Meta-tools do not reduce the amount of tokens per step, but remove whole steps, bypassing the output token cost for that step.

\begin{figure}
    \centering
    \includegraphics[width=0.9\linewidth]{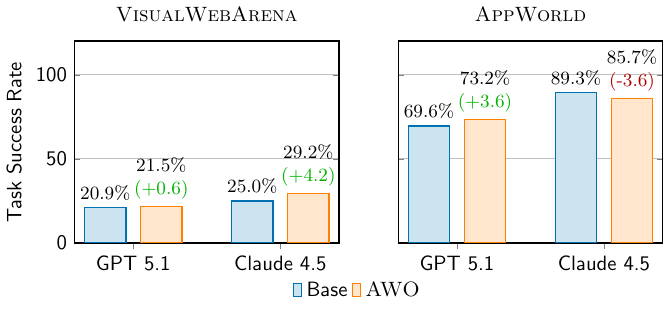}
    \caption{The task success rate with and without \sys, when using the GPT 5.1 and Claude 4.5 models, for \appworld (left) and \visualwebarena (right).}
    \label{fig:pass_rate}
\end{figure}

\textbf{Task success rate.}
\Cref{fig:pass_rate} shows the task success rate across models and benchmarks, with and without \sys.
In general, \sys contributes positively to the task success rate, which we explain as follows.
On one hand, meta-tools decrease the average number of steps to completion, \eg, when using Claude 4.5, \sys reduces this number from 16.7 to 15.4 on \appworld and from 10.8 to 9.7 on \visualwebarena.
This reduces the agent context size within a single execution which in turn reduces potential errors.
On the other hand, meta-tools allow agents to quickly complete overarching sub-tasks that would otherwise necessitate reasoning over and invoking multiple fine-grained tools.
We see a slight decrease in task success rate for \sys on \appworld with Claude 4.5.
However, we manually verify that in this setting the created meta-tools are appropriate and we attribute this decrease to randomness by the \ac{LLM} process when running tasks that are not using meta-tools.

\subsection{Meta-tools Analysis}

We now analyze the effect of meta-tools on the execution trajectories across both benchmarks.

\textbf{Meta-tool Utilization.}
The utilization rate of meta-tools varied significantly between benchmarks and models.
In \appworld, meta-tool utilization was remarkably high: both GPT 5.1 and Claude 4.5 achieved 98.2\% utilization, indicating that nearly all tasks benefited from the auto-login meta-tools.
In \visualwebarena, utilization was more moderate: 30.6\% and 16.3\% for the GPT 5.1 and Claude 4.5 models, respectively.
This difference reflects the nature of the benchmarks, \appworld tasks consistently require authentication across its simulated applications, making login meta-tools universally applicable, while \visualwebarena's tasks split earlier, making the meta-tools inherently more specialized.

\begin{figure}
    \centering
    \includegraphics[width=0.7\linewidth]{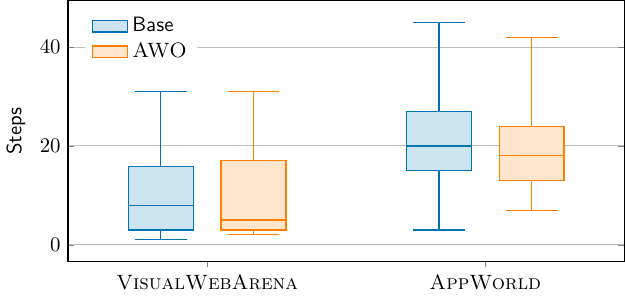}
    \caption{The distribution of steps for tasks that used Meta-Tools with and without \sys using GPT 5.1 for \visualwebarena and \appworld.}
    \label{fig:meta_step_analysis}
\end{figure}

\textbf{Meta-tools impact.}
\Cref{fig:meta_step_analysis} shows the distribution of the number of steps for tasks that used meta-tools comparing them to the same tasks with the base toolset with the GPT 5.1 model (the full distribution is available in Appendix~\ref{app:appendix_results}).
On average, tasks using meta-tools take 1.0 less steps on \visualwebarena and 2.63 less steps on \appworld.
This result highlights that when they are used, meta-tools effectively reduce the length of trajectories.

\subsection{Merging Analysis}

The number of meta-tools we can extract is linked to the level of horizontal merging we can achieve.
\Cref{tab:combined_compression_results} displays the effect of horizontal merging on the state graphs of \visualwebarena and \appworld.
Importantly, while horizontal merging provides multiple candidates for meta-tools, they might not be realizable in cases, where the tools are of high complexity or have a wide input space.
Thus, we observe that while the merging level on \visualwebarena ($-16.5\%$) is in the same range as the final LLM steps saved ($-10.2\%$), \appworld merging did not follow this trend.
This is due to many candidate meta-tools in \appworld not being reasonably convertable.

\begin{table}[]
    \centering
    \caption{Comparison of node, edge, and sink counts before and after horizontal merging for \visualwebarena and \appworld.}
    \small
    \setlength{\tabcolsep}{8pt} %
    \begin{tabular}{l ccc}
        \toprule
        \textbf{Strategy} & \textbf{$|V|$} & \textbf{$|E|$} & \textbf{Sinks} \\
        \midrule
        \multicolumn{4}{l}{\visualwebarena} \\
        State graph & 6216 & 6213 & 870 \\
        Merged state graph & 5189 ($-16.5\%$) & 5186 & 863 \\
        \midrule
        \multicolumn{4}{l}{\appworld} \\
        State graph & 2427 & 2427 & 82 \\
        Merged state graph   & 249 ($-89.7\%$) & 447  & 47 \\
        \bottomrule
        \bottomrule
    \end{tabular}
    \label{tab:combined_compression_results}
\end{table}

\section{Related Work}
\label{sec:related}

\textbf{Agentic Infrastructure Optimizations.}
Recent work has explored various approaches to improve the execution efficiency of agentic systems.
\textsc{LLMCompiler}~\cite{kim2024llm} achieves latency speedups and cost savings through parallel tool execution via DAG-based planning, identifying independent tool calls that can be executed concurrently.
\textsc{ToolChain*}~\cite{zhuang2024toolchain} formulates tool-use planning as A* search over a decision tree of potential API calls, reducing exploration time while improving accuracy.
\textsc{LLM-Tool Compiler}~\cite{singh2024llm} proposes runtime tool fusion inspired by hardware multiply-add operations, combining related calls dynamically during execution.
\textsc{ADAS}~\cite{hu2025automated} introduces meta-agents that automatically design agent architectures through code generation.
Unlike methods that optimize runtime execution or tool composition, \sys statically extracts meta-tools from trajectories before deployment.
This eliminates runtime overhead and ensures deterministic sub-task execution.
These approaches are complementary: \sys’s meta-tools can be further enhanced via parallel execution or improved search strategies.

\textbf{Tool Selection and Prompting.}
A separate line of work targets efficiency through improved reasoning and tool selection strategies.
\react~\cite{yao2023react} interleaves reasoning traces with actions, enabling agents to dynamically plan and adjust tool usage based on intermediate observations.
\textsc{Tree of Thought}~\cite{yao2024tree} extends chain-of-thought prompting by exploring multiple reasoning paths, allowing for deliberate planning and backtracking during complex problem-solving.
\textsc{AVATAR}~\cite{avatar2024} employs contrastive reasoning between successful and failed trajectories to teach agents better tool selection, however, this requires additional training on trajectory data.
Recent work on joint optimization~\cite{wu2025joint} focuses on refining agent prompts and tool descriptions to reduce unnecessary calls.
\textsc{ACON}~\cite{kang2025acon} compresses environment observations and interaction histories, reducing peak token usage while preserving task accuracy.
Beyond improving reasoning or selection, \sys reduces the agent's decision load by encapsulating multi-step behaviors into single, higher-level abstractions.

\textbf{LLM Optimizations.}
Orthogonal to tool-level optimizations, some works have focused on reducing inference costs at the model level.
Techniques such as KV-cache optimizations~\cite{kwon2023efficient,pope2023efficiently,adnan2024keyformer} and \textsc{FlashAttention}~\cite{dao2022flashattention,dao2023flashattention,shah2024flashattention} reduce computational overhead during generation.
While these approaches optimize token and memory efficiency, \sys provides a complementary benefit by reducing total \ac{LLM} invocations through meta-tools, allowing for their combined application.

\textbf{Code Optimizations.}
The core idea behind \sys is similar to the idea of function inlining in compilers and kernel fusion in GPU-programming.
Function inlining has been extensively studied as a means to eliminate call overhead and enable further optimizations~\cite{davidson1988study, chang1992profile, chakrabarti2006inline, theodoridis2022understanding}.
Profile-guided approaches use execution traces to identify candidates for inlining~\cite{chang1992profile}, analogous to how \sys analyzes agent trajectories to identify common tool sequences.
Kernel fusion in GPU computing~\cite{wang2010kernel, fousek2011automatic, wahib2014scalable} combines multiple operations into single kernels to reduce memory traffic and improve efficiency, a concept parallel to our meta-tool abstraction.
We apply similar principles to agentic workflows by identifying common tool sequences within stochastic, \ac{LLM}-generated execution traces and fusing them into meta-tools.
This adapts traditional compiler optimization to non-deterministic control flows.

\section{Conclusions}
\label{sec:conclusion}

We introduced \sys, a framework that analyzes agentic workflows to compile recurring agent behaviors into deterministic meta-tools.
\sys reduces operational cost and end-to-end latency of running agentic AI while improving task success by replacing redundant \ac{LLM} reasoning with reusable composite actions provided by meta-tools.
Our experimental evaluation on two benchmarks suggest that \sys is a practical path toward making agentic systems more efficient and reliable in production settings.

\section*{Impact Statement}
This paper presents work whose goal is to advance the field of machine learning systems.
There are many potential societal consequences of our work, none of which we feel must be specifically highlighted here.

\section*{LLM Usage Statement}
We acknowledge the use of various LLM assistants to help retrieve information such as related work, baselines, and help polish the writing of the paper.
However, all ideas, designs, and writing were developed and verified by the authors.

\section*{Acknowledgment}
This work has been co-funded by the Swiss National Science Foundation, under the project FRIDAY: Frugal, Privacy-Aware and Practical Decentralized Learning, SNSF proposal No. 10.001.796.

\bibliography{main.bib}
\bibliographystyle{icml2026}

\clearpage

\appendix

\section{Visual Web Arena}
\label{app:vwa_metatools}

\subsection{Horizontal Merging Rules}

To identify common path in the trajectories, we implement compression rules to map semantically equivalent states in the state graph to the same nodes.
In Visual Web Arena, we do this by identifying ids, used by the agent for its interactions with the web pages, that corresponds to the same element.

\paragraph{State Definition}
As Visual Web Arena relies on a Tool Calling Agent, every LLM reasoning call leads to exactly one action.
Thus, each node of the state graph can be the tool call generated by the LLM call.
As actions are different when applied to a page with a different url (for example, click on id 2 for two different pages lead to two different results), the state is defined as the tuple (tool\_call, current\_url).

The horizontal merging applied to this state definition consists in rewriting rules mapping semantically equivalent tuple to the same tuple.
We now present these rules for each suite of the benchmark.

\paragraph{Reddit}
The reddit website has a navbar available on every pages.
In particular, the element with id 3 is always the search bar and the element with id 4 is the search button.
The first rewriting rules map the \texttt{type [3]} to \texttt{type\_search [string]} and \texttt{click [4]} to \texttt{start\_search}.
Note that including a line break in the search string directly starts the search.
In such cases, a \texttt{start\_search} is added to allow identification during the Meta-Tool extraction.

The second main rewriting rule focuses on post comments.
Every post includes a comment textarea.
Using the html of each specific page, we identify the id of the text area and map \texttt{type [comment\_area\_id]} to \texttt{write\_comment [string]}.
We also identify the post button id to map \texttt{click [post\_id]} to \texttt{post\_comment}.
Note that all type actions happening before clicking the post button are always concatenated into a single \texttt{write\_comment}.
Finally, we map fixed elements of the navbar such as buttons redirecting to the forum list or to the user account to their specific actions.

\paragraph{Classifieds}
Similarly to reddit, the classifieds website has a search bar.
This search bar is only present on the main page and the category pages and at different positions using different interaction ids.
We identify the id for each url and map \texttt{type [search\_id]} to \texttt{type\_search [string]} and \texttt{click [search\_button]} to \texttt{start\_search}.
We make sure to add a \texttt{start\_search} to type actions including a line break.

Like the \texttt{post\_comment} action in reddit, we identify the comment title and body text areas, and map the type actions accordingly, making sure to merge several type actions into a single \texttt{write\_title} or \texttt{write\_body}.

\paragraph{Shopping}
The shopping website has a navbar with a search bar always available.
The first mapping rule maps any type to this bar into a \texttt{write\_search [string]} and clicks to \texttt{start\_search}.
Notably, because of the website layout, the id of the search bar can be different and we identify it for each url to ensure correct identification of the \texttt{type} states.

Similarly to the post comments in both previous websites, shopping has a review features.
We identify the id of the title and review text areas and map type actions to \texttt{write\_title} and \texttt{write\_review}.
Again, all types to these elements are merged into a single action.
Reviews also include a rating system requiring the user to click on the number of stars corresponding to the rating.
We identify the id of each start and map them to \texttt{set\_rating [rating]} states.

\subsection{Visual Web Areana Meta-Tools}

We list the meta-tools created for each suite in Visual Web Arenas with their associated tool calls in \Cref{tab:vwa_mt}.
\begin{table*}[t] %
\centering
\caption{Implemented meta-tools in each suite of \visualwebarena and their corresponding tool calls.}
\begin{tabular}{|l|l|l|}
\hline
Website & Meta-tool & Tool Call\\
\hline
\multirow{2}{*}{Reddit} & search {[}string{]} &
    \begin{tabular}[c]
    {@{}l@{}}type {[}3{]} {[}string{]}\\
    click {[}4{]}
    \end{tabular}\\
\cline{2-3}
& post\_comment {[}comment{]} &
    \begin{tabular}[c]
    {@{}l@{}}scroll //
    scroll to comment box\\
    type {[}comment\_box\_id{]} {[}comment{]}\\
    click {[}post\_comment\_id{]}
    \end{tabular}\\
\hline
\multirow{2}{*}{Classifieds} & search {[}string{]} &
    \begin{tabular}[c]
    {@{}l@{}}type {[}search\_box\_id{]} {[}string{]}\\
    click {[}search\_submit\_id{]}
    \end{tabular}\\
\cline{2-3}
& post\_comment {[}title{]} {[}comment{]} &
    \begin{tabular}[c]
    {@{}l@{}}scroll // scroll to comment box\\
    type {[}title\_box\_id{]} {[}title{]}\\
    type {[}comment\_box\_id{]} {[}comment{]}\\
    click {[}post\_comment\_id{]}
    \end{tabular}\\
\hline
\multirow{2}{*}{Shopping} & search {[}string{]} &
    \begin{tabular}[c]
    {@{}l@{}}type {[}search\_box\_id{]} {[}string{]}\\
    click {[}search\_submit\_id{]}
    \end{tabular}\\
\cline{2-3}
& leave\_review {[}rating{]} {[}title{]} {[}review{]} &
    \begin{tabular}[c]
    {@{}l@{}}click {[}review\_tab\_id{]}\\
    scroll // scroll to review box\\
    click {[}rating\_star{]}\\
    type {[}title\_box\_id{]} {[}title{]}\\
    type {[}review\_box\_id{]} {[}review{]}\\
    click {[}post\_review\_id{]}
    \end{tabular}\\
\hline
\end{tabular}
\label{tab:vwa_mt}
\end{table*}

\section{Appworld methods}
\label{app:appworld_metatools}

\subsection{State graph for coding agents}

In a coding agent, the agent can output multiple toolcalls in one trip to the LLM engine.
Thus, theoretically, with a perfect agent and context engineering, the agent could solve a whole task with one output, assuming the environment outputs don't change toolcalls used.
Thus, using tool calls to define states is a proxy and merging them only indirectly reduces the number of LLM calls, as they get access to more direct tools.
However, as coding agents often take small steps, this still reduces the amount of times an agent uses its LLM.
This is backed by our positive results in Section~\ref{sec:results}.

\subsection{Merging Heuristics for AppWorld}
\label{app:compression_heuristics}

In order to extract popular execution paths (hot paths) from a state graph, we must merge them.
In Appworld, we do this through relaxing the state rule, applying regex rules to the execution logs, making domains commutative and merging recurrent actions.

\paragraph{State Definition}
In a merged state graph for Appworld, a node is defined by

$$V = \text{Hash}( \Phi_{GET} \oplus \Phi_{ACTION} )$$

$\Phi_{GET}$ is a \textbf{set} of all gathering actions the agent has done and $\Phi_{ACTION}$ is a \textbf{list} of all other actions. This distiction is made to make all read-like actions commutative and idempotent, for example merging workflows, where agents read the same documents in a different order. If the agent reads the same document twice, it does not change the state, as read actions are recorded in a set.

\paragraph{Regex Merging}
Regex rules are used mostly to remove the relevance of specific arguments from the execution logs. An example of a regex rule would be mapping \texttt{user\_id=\textbackslash d+} $\to$ \texttt{user\_id=\{UID\}} in all tasks, so as to merge the same task done for different users.

\paragraph{Domain Commutativity}
Standard APIs often consist of functionally independent services (e.g., Gmail vs. Spotify). We define a rule where actions in disjoint domains are treated as commutative.
$$ \text{State}(A_{gmail} \to B_{spotify}) \equiv \text{State}(B_{spotify} \to A_{gmail}) $$
This heuristic collapses "diamond" structures in the graph, merging diverging paths that achieve the same multi-domain outcome.

\paragraph{Recurrent Action Merging}
Agents often perform repetitive actions, such as paging through search results. We detect these repetitions and compress them into a single self-referential loop node ($A \circlearrowleft$) to prevent the graph from expanding infinitely during long scraping tasks.

\begin{table}[h]
    \centering
    \caption{Graph topology statistics through different compression strategies. $|V|$: Nodes, $|E|$: Edges, $\overline{d_{in}}$: Average In-Degree. Strategies are cumulative top to bottom. Graphs constructed from 82 GPT-4o traces containing 2427 total API calls.}
    \setlength{\tabcolsep}{5pt}
    \begin{tabular}{l ccccc}
        \toprule
        \textbf{Strategy} & \textbf{$|V|$} & \textbf{$|E|$} & \textbf{Sinks} & \textbf{$\overline{d_{in}}$} & \textbf{Endpts} \\
        \midrule
        Disjoint (Base) & 2428 & 2427 & 82 & 1.00 & 2427 \\
        + GETs          & 1211 & 1437 & 73 & 1.19 & 638 \\
        + Regex         & 750  & 968  & 66 & 1.29 & 331 \\
        + Actions       & 249  & 447  & \textbf{47} & 1.80 & 77 \\
        \bottomrule
    \end{tabular}
    \label{tab:appworld_compression_levels}
\end{table}

Importantly, our collection of merging rules is not exhaustive, as for any agentic system there might be more rules to be discovered by the expert human or agent. The resulting graph statistics through a cumulative application of merging strategies on the \appworld benchmark can be seen in Table~\ref{tab:appworld_compression_levels}. The detected meta-tools for AppWorld using these heuristics are found in Table~\ref{tab:aw_mt}.

\begin{table*}[t]
\centering
\caption{Implemented meta-tools in \appworld and the multi-step API calls they replace.}
\begin{tabular}{|l|l|}
\hline
 Meta-Tool & Replaced Tool Call Sequence\\
\hline
 spotify\_auto\_login() &
    \begin{tabular}[c]
    {@{}l@{}}supervisor.show\_account\_passwords()\\
    spotify.login(username, password)
    \end{tabular}\\
\hline
 file\_system\_auto\_login() &
    \begin{tabular}[c]
    {@{}l@{}}supervisor.show\_account\_passwords()\\
    file\_system.login(username, password)
    \end{tabular}\\
\hline
 venmo\_auto\_login() &
    \begin{tabular}[c]
    {@{}l@{}}supervisor.show\_account\_passwords()\\
    venmo.login(username, password)
    \end{tabular}\\
\hline
 phone\_auto\_login() &
    \begin{tabular}[c]
    {@{}l@{}}supervisor.show\_account\_passwords()\\
    phone.login(username, password)
    \end{tabular}\\
\hline
 simple\_note\_auto\_login() &
    \begin{tabular}[c]
    {@{}l@{}}supervisor.show\_account\_passwords()\\
    simple\_note.login(username, password)
    \end{tabular}\\
\hline
\end{tabular}
\label{tab:aw_mt}
\end{table*}

\section{Automated Optimization Loop}
\label{sec:automated_loop}

While human analysis is effective for identifying initial meta-tool candidates, scaling this process requires automation. We explored whether an agent could autonomously discover merging rules by directly analyzing execution traces. To this end, we implemented an agentic loop that iteratively sweeps over logs and defines regex rules for trace merging.

\subsection{Optimization Primitives}
The optimization agent operates through a set of three discrete actions. They allow the agent to modify the graph topology by identifying semantic equivalences and removing redundancy. The actions mirror the merging heuristics described in Appendix~\ref{app:compression_heuristics}.

\paragraph{regex\_sub}
Identifies unimportant noise within execution traces (e.g., timestamps, ephemeral session IDs) and replaces them with static placeholders (e.g., mapping \texttt{user\_id=8372} $\to$ \texttt{user\_id=\{UID\}}). \\
This action forces distinct execution paths to collapse into a single path.

\paragraph{set\_domain}
Tags operations with specific functional domains (e.g., \texttt{SPOTIFY}, \texttt{GMAIL}). The graph construction algorithm treats actions in disjoint domains as commutative. This action allows the graph to merge diverging "diamond" structures (where $A \to B$ vs $B \to A$) into a single unified flow.

\paragraph{set\_semantic\_type}
Categorizes operations based on their side-effect profile:
\begin{itemize}
    \item \textbf{Annotator:} Internal logging or reasoning (Zero state change).
    \item \textbf{Accessor:} Read-only operations (Idempotent).
    \item \textbf{Mutator:} Write operations (State-changing).
\end{itemize}
Annotators are excluded from the state hash, and Accessors (readers) only affect the state on their first usage, debloating the graph.

\subsection{The Optimization Cycle}
\begin{figure}[h]
    \centering
    \includegraphics[width=0.4\linewidth]{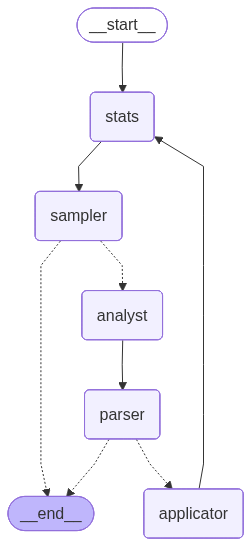}
    \caption{Horizontal merger agent structure. The agent has the capability to calculate stats, visualise logs, decide merging rules and apply them}
    \label{fig:compression_agent}
\end{figure}

The discovery of these rules is driven by an iterative agentic loop, illustrated in Figure \ref{fig:compression_agent}. The process repeats for a maximum of $N$ iterations or until the agent finds no new rules.

The workflow consists of five distinct stages:
\begin{enumerate}
    \item \textbf{Stats:} Calculates graph complexity metrics (node/edge count, branching factor) to quantify progress.
    \item \textbf{Sampler:} Selects a random subset of traces to not overflow the LLM with too many traces.
    \item \textbf{Analyst:} The LLM node. It scans the sample to discover patterns and propose new merge rules (e.g., \texttt{regex\_sub}).
    \item \textbf{Parser:} Validates the LLM output structural correctness. If no valid rules are found, the system routes back to the sampler.
    \item \textbf{Applicator:} Executes validated rules against the \textit{entire} global dataset, permanently transforming the trace graph for the next cycle.
\end{enumerate}

\subsection{Agentic merging results}

The results of the agentic merger are preliminary, as we do not yet have a proper methodology for evaluating the quality of its work. As the number of regex rules output could reach thousands, verifying their correctness or setting guardrails could be more difficult, then just having a domain expert do the analysis themselves.

The first tests indicate it is capable of large merges, with the caveat that it hasn't been able to merge full paths like a human has been.

\begin{table}[h]
    \centering
    \caption{Graph topology statistics across different optimization iterations. Data shows the progression of one graph merging execution from Iteration 0 (disjoint graph) to 100.}
    \setlength{\tabcolsep}{5pt}
    \begin{tabular}{l ccccc}
        \toprule
        \textbf{Iteration} & \textbf{$|V|$} & \textbf{$|E|$} & \textbf{Sinks} & \textbf{$\overline{d_{in}}$} & \textbf{Endpts} \\
        \midrule
        0   & 5432 & 5443 & 129 & 1.00 & 4170 \\
        1   & 4166 & 4403 & 129 & 1.06 & 3689 \\
        2   & 3943 & 4199 & 128 & 1.06 & 3425 \\
        3   & 2013 & 2408 & 128 & 1.20 & 1647 \\
        15  & 1490 & 1822 & 116 & 1.22 & 965 \\
        50  & 1477 & 1806 & 113 & 1.22 & 959 \\
        100 & 1163 & 1423 & \textbf{97} & 1.22 & 821 \\
        \bottomrule
    \end{tabular}
    \label{tab:graph_stats_iterations}
\end{table}

Table~\ref{tab:graph_stats_iterations} displays the merged state graph statistics through 100 merging-agent iterations. We observe that on 168 Appworld tasks it was able to reduce the number of nodes from 5432 to 1163.

\begin{figure}[h]
    \centering
    \includegraphics[width=0.9\linewidth]{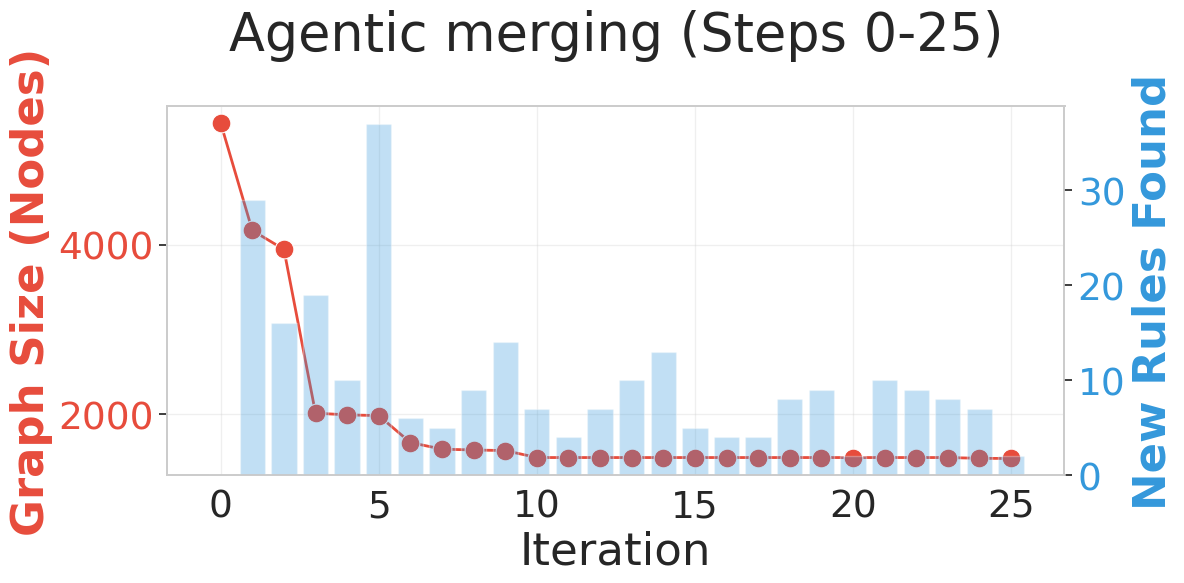}
    \caption{Graph compression through iterations of the agent cycle in Figure\ref{fig:compression_agent}. The Merging is very effective at the start and even though the number of rules output does not change drastically, the agent is unable to achieve merging levels a human would}
    \label{fig:agent_compression_plot}
\end{figure}

Figure~\ref{fig:agent_compression_plot} shows the number of rules generated by the agent compared to the level of merging the rules achieved through 25 iterations of the merging agent. Notably, the compression is very effective at the start, but quickly stagnates, as the agent struggles to discover meaningful rules, that actually have an effect on the horizontal merging process.

Furthermore, due to the agent's volatility and sheer number of rules it outputs, it is difficult to evaluate whether all of the rules it created were good, or whether some of them were bad enough to contaminate the whole system, making meta-tool discovery a gamble. Verification of an agent rule and meta-tool discoverer remains a future challenge.

\section{Additional Results}
\label{app:appendix_results}

In addition to GPT 5.1 and Claude 4.5, we ran our experiments with an open source model GPT-OSS. The full quantitative results of the three models are recorded in Table~\ref{tab:vwa_results} for \visualwebarena and in Table~\ref{tab:aw_results_appendix} for Appworld.
Note that a breakdown of all results with the three suites available in \visualwebarena is available in \Cref{tab:vwa_per_dataset}.
They display the accuracy, number of LLM steps, total cost, latency, meta-tool utilisation and the number of interactions with the environment through API calls and database operations through an execution with an agent that either has or does not have access to meta-tools. We observe that nearly all cost-related metrics reduce as a result of introducing meta-tools.

However, interestingly, while GPT-OSS benefitted from having meta-tools in terms of task completion, it became less efficient at both benchmarks. We empirically observed that this is due to it often forgetting what it had done. Therefore it often picked using meta-tools as the path forward, since they leave a powerful impression.

Tables~\ref{tab:token_cost_embedded_vwa} and \ref{tab:token_cost_embedded_aw} break down the total token usage of running the full benchmarks before and after adding meta-tools.
It can be seen, that the most significant cost comes from output tokens.
Since the agent's memory largely remains the same through an execution, the cost from cache hits is not great.
This means that reducing the number of times the \ac{LLM} has to generate new toolcalls is the most expensive part of the procedure, which is why we prioritize skimming down on the trips to the \ac{LLM} altogether

\begin{table*}[h]
\centering
\caption{Results for the \visualwebarena benchmark. $^\dagger$GPT-OSS results are for the Reddit dataset only using only the textual data.}

\small
\begin{tabular}{|l|rr|rr|rr|}
\hline
 & \multicolumn{2}{c|}{\textbf{GPT 5.1}} & \multicolumn{2}{c|}{\textbf{Claude 4.5}} & \multicolumn{2}{c|}{\textbf{GPT-OSS}$^\dagger$} \\
 & Base & \sys & Base & \sys & Base & \sys \\
\hline
Task success rate & 20.9\% & 21.5\% & 25.0\% & 29.2\% & 6.1\% & 10.9\% \\
Total LLM call count & 9433 & 8906 (--5.59\%) & 8695 & 7807 (--10.21\%) & 2141 & 2548 (+19.01\%) \\
E2E latency (hours) & 52.87 & 53.00 & 40.17 & 36.55 & 12.84 & 15.15 \\
Steps/task & 11.2 & 10.6 & 10.8 & 9.7 & 13.0 & 15.4 \\
Task duration (average in seconds) & 226.0 & 226.6 & 178.98 & 162.82 & 280.19 & 330.65 \\
Meta-tool utilisation & --- & 30.6\% & --- & 16.3\% & --- & 31.5\% \\
\hline
\end{tabular}
\label{tab:vwa_results}
\end{table*}

\begin{table*}[h]
\centering
\caption{Per-dataset breakdown for the \visualwebarena benchmark.}
\small
\begin{tabular}{|l|l| cccc|}
\hline
\textbf{Dataset} & \textbf{Model} & \textbf{Task success rate} & \textbf{Total Steps} & \textbf{Duration (hrs)} & \textbf{Meta-tool Usage} \\
\hline
Reddit & GPT 5.1 & 17.6\% & 3440 & 17.56 & --- \\
       & GPT 5.1 w/ \sys & 18.5\% & 3080 & 16.41 & 33.2\% \\
       & Claude 4.5 & 15.3\% & 2313 & 9.28 & --- \\
       & Claude 4.5 w/ \sys & 19.8\% & 2098 & 10.89 & 20.3\% \\
\hline
Shopping & GPT 5.1 & 20.9\% & 3532 & 27.17 & --- \\
         & GPT 5.1 w/ \sys & 22.3\% & 3399 & 24.87 & 35.2\% \\
         & Claude 4.5 & 27.8\% & 3863 & 21.35 & --- \\
         & Claude 4.5 w/ \sys & 32.3\% & 3453 & 18.89 & 14.8\% \\
\hline
Classifieds & GPT 5.1 & 24.5\% & 2461 & 8.14 & --- \\
            & GPT 5.1 w/ \sys & 22.9\% & 2427 & 11.72 & 17.0\% \\
            & Claude 4.5 & 27.6\% & 2519 & 9.54 & --- \\
            & Claude 4.5 w/ \sys & 30.7\% & 2256 & 6.77 & 16.1\% \\
\hline
\end{tabular}
\label{tab:vwa_per_dataset}
\end{table*}

\begin{table*}[]
    \centering
    \caption{Results for the \appworld benchmark.}
    \begin{tabular}{|l|cc|cc|cc|}
        \hline
        \multirow{2}{*}{} & \multicolumn{2}{c|}{\textbf{GPT 5.1}}  & \multicolumn{2}{c|}{\textbf{Claude 4.5}} & \multicolumn{2}{c|}{\textbf{GPT-OSS}} \\
        & Base & \sys & Base & \sys & Base & \sys \\
        \hline
        Task success rate & $69.6\%$ & $73.2\%$ & $89.3\%$ & $85.7\%$ & 14.3\%&  16.1\%\\
        Total LLM call count & 3665 & 3229 ($-11.90\%$)& 2801& 2588 ($-7.60\%$)& 2137& 2226 ($+4.20\%$)\\
        E2E latency (hours) & 12.44 & 9.37& 3.53& 4.11& 3.32& 3.49 \\
        LLM call/task & 21.8 & 19.2& 16.7& 15.4& 13.2 & 12.7\\
        task duration (average in seconds) & 266.6 & 200.8& 75.6& 88.1& 71.1& 74.8 \\
        Meta-tool utilisation & & 98.2\%& & 98.2\%& & 39.3\%\\
        total API calls & 9115& 6415& 5010& 4869& 6253& 8821\\
        API calls/task & 54.3& 38.2& 29.8& 29.0& 37.2& 52.5\\
        DB ops/task & 26.2& 25.7& 25.8& 25.5& 25.4& 30.5\\
        \hline
    \end{tabular}
    \label{tab:aw_results_appendix}
\end{table*}

\begin{table*}[h]
    \centering
    \caption{Token usage and cost breakdown for the \visualwebarena benchmars. Prices are shown per 1 million tokens. Note that GPT does not incur a separate cache write fee.}
    \setlength{\tabcolsep}{5pt}
    \begin{tabular}{|l|c|cc|c|cc|}
        \hline
        \multirow{2}{*}{\textbf{Metric}} & \multicolumn{3}{c|}{\textbf{GPT 5.1}} & \multicolumn{3}{c|}{\textbf{Claude 4.5}} \\
         & \textit{Price/1M} & Base & \sys & \textit{Price/1M} & Base & \sys \\
        \hline
        Input Cache Miss & \$1.25 & 25.5M & 24.1M & \$3.00 & 25.8M & 23.2M \\
        Input Cache Hit & \$0.125 & 68.8M & 65.9M & \$0.30 & 50.6M & 45.4M \\
        Input Cache Write & — & 0 & 0 & \$3.75 & 42.4M & 38.1M \\
        Output Tokens & \$10.00 & 0.39M & 0.36M & \$15.00 & 1.4M & 1.2MM \\
        \hline
        \textbf{Total Tokens} & & \textbf{95.7M} & \textbf{90.3M} & & \textbf{120.1M} & \textbf{108.0M} \\
        \textbf{Total Cost (\$)} & & \textbf{44.5} & \textbf{42.0} & & \textbf{272.3} & \textbf{244.5} \\
        \hline
    \end{tabular}
    \label{tab:token_cost_embedded_vwa}
\end{table*}

\begin{table*}[h]
    \centering
    \caption{Token usage and cost breakdown for the \appworld benchmark (168 tasks). Prices are shown per 1 million tokens. Note that GPT does not incur a separate cache write fee.}
    \setlength{\tabcolsep}{5pt}
    \begin{tabular}{|l|c|cc|c|cc|}
        \hline
        \multirow{2}{*}{\textbf{Metric}} & \multicolumn{3}{c|}{\textbf{GPT 5.1}} & \multicolumn{3}{c|}{\textbf{Claude 4.5}} \\
         & \textit{Price/1M} & Base & \sys & \textit{Price/1M} & Base & \sys \\
        \hline
        Input Cache Miss & \$1.25 & 3.69M & 2.99M & \$3.00 & 2.21M & 2.19M \\
        Input Cache Hit & \$0.125 & 28.04M & 23.96M & \$0.30 & 24.92M & 22.17M \\
        Input Cache Write & — & 0 & 0 & \$3.75 & 2.21M & 2.18M \\
        Output Tokens & \$10.00 & 2.35M & 2.01M & \$15.00 & 0.48M & 0.47M \\
        \hline
        \textbf{Total Tokens} & & \textbf{34.08M} & \textbf{28.97M} & & \textbf{29.82M} & \textbf{27.01M} \\
        \textbf{Total Cost (\$)} & & \textbf{31.59} & \textbf{26.85} & & \textbf{29.63} & \textbf{28.39} \\
        \hline
    \end{tabular}
    \label{tab:token_cost_embedded_aw}
\end{table*}

\end{document}